\documentclass[unnumsec,webpdf,contemporary,large]{oup-authoring-template}%

\graphicspath{{Fig/}}

\theoremstyle{thmstyleone}%
\theoremstyle{thmstyletwo}%
\theoremstyle{thmstylethree}%

\begin{document}

\journaltitle{Journal Title Here}
\DOI{DOI added during production}
\copyrightyear{YEAR}
\pubyear{YEAR}
\vol{XX}
\issue{x}
\access{Published: Date added during production}
\appnotes{Paper}

\firstpage{1}


\title[Short Article Title]{SYNRARE: Synthetic Rare Disease EHR Generation for ML Benchmarking}

\author[1]{Nicolai Dinh Khang Truong\ORCID{0009-0002-8622-5237}}
\author[1,$\ast$]{Richard Röttger\ORCID{0000-0003-4490-5947}}

\address[1]{\orgdiv{Department of Mathematics and Computer Science}, \orgname{University of Southern Denmark}, \orgaddress{\street{Campusvej 55}, \postcode{5230}, \state{Odense}, \country{Denmark}}}

\corresp[$\ast$]{Corresponding author. \orgdiv{Department of Mathematics and Computer Science}, \orgname{University of Southern Denmark}, \orgaddress{\street{Campusvej 55}, \postcode{5230}, \state{Odense}, \country{Denmark.}} E-mail: \href{email:roettger@imada.sdu.dk}{roettger@imada.sdu.dk}}

\received{Date}{0}{Year}
\revised{Date}{0}{Year}
\accepted{Date}{0}{Year}



\abstract{\textbf{Motivation:} Rare disease (RD) diagnosis is frequently delayed due to the similarities in symptoms to common disease variants. Machine Learning Algorithms applied to Electronic Health Records show promise for accelerating the diagnosis; however, legal and privacy concerns pose significant barriers. To address these issues, Synthetic Data Generation is an alternative method for obtaining Electronic Health Records and can be applied with any Machine Learning algorithm for benchmarking and development purposes. Despite the availability of Synthetic Data Generation algorithms, support for generating a subset of patients that differ in a definable degree from the majority to simulate patients with RD is often lacking. \\
\textbf{Results:} We present SYNRARE, a graphical user interface based on the Synthea framework that enables easier modification and generation of synthetic Electronic Health Records of RD patients, which differ only to a definable degree from patients with common diseases, thereby enabling the benchmarking and testing of algorithms under controlled technical conditions. SYNRARE enables researchers to rapidly benchmark their Machine Learning algorithms across any scenario. \\
\textbf{Availability and implementation:} SYNRARE, including detailed instructions for installing, is available at https://gitlab.sdu.dk/screen4care/synrare. \\} 



\maketitle


\section{Introduction}
Rare diseases are defined as conditions affecting 1 in 2,000 patients in the European Union \citep{Valdez2016PublicHA, Health2024TheLF}.
Due to the low prevalence of these diseases, their symptomatological similarity to common diseases, and clinical unfamiliarity, patients undergo a diagnostic odyssey, taking several years to receive the final diagnosis \citep{Valdez2016PublicHA, Health2024TheLF, dubief2024newborn}.
Machine Learning on Electronic Health Records (EHRs) can accelerate the diagnosis of patients with rare diseases and therefore help shorten the path to diagnosis.
Recent examples demonstrate that applying Machine Learning (ML) algorithms alone to EHRs can identify rare diseases, thereby providing clinicians with the information necessary to perform targeted interventions \citep{Huda2021AML, Cohen2020DetectingRD, Wilson2023DevelopmentOA}.
However, rare diseases generally pose a challenge to ML approaches: We are normally presented with highly imbalanced datasets, resulting in increased false positives due to Bayes' theorem. Tailored and specialized ML approaches have to be developed and tested in controlled environments to demonstrate their capabilities for the application in those cases. 
However, due to privacy and ethical concerns \citep{Fecho2022LeveragingOE}, access to extensive real-world data is limited and consequently poses a significant roadblock for research on the limitations and suitability of novel ML methods for EHR data of patients with rare diseases.
Synthetic data helps overcome these barriers, allowing researchers to conduct rapid benchmarking and evolution of the capabilities of their ML models in a controlled environment.
For example, a research group can test the ability of a classifier or outlier detection method for the early detection of rare diseases and train and evaluate it on synthetic patient data under realistic circumstances before undergoing the diligent access procedure to finally access real-world data.

We present SYNRARE, a GUI that leverages the Synthea framework \citep{walonoski2018synthea}, enabling researchers to quickly modify existing modules of common diseases to generate synthetic EHRs, including examples of slightly modified patients with a simulated rare disease. 
SYNRARE is readily available at https://gitlab.sdu.dk/screen4care/synrare, thereby enabling researchers to quickly set up a testbed for technical ML evaluation. It is important to note that SYNRARE is intended for the technical evaluation of ML procedures, where researcher seek to evaluate the technical capabilities of their algorithm to differentiate a minority class from a majority class with realistically appearing EHR data. 
Furthermore, we encourage users, especially clinical experts, to contribute to the project to capture accurate disease progression. While SYNRARE allows for the inclusion of literature-based knowledge, the data is not intended to help understand the rare disease in question, but merely to create cohorts that have a definable degree of dissimilarity from the majority class.
Synthea is an open-source, rule-based synthetic data generator for patients that simulates disease trajectories for each patient from birth to the present.
The framework consists of selectable disease-specific modules designed and developed by medical experts.
We choose Synthea because it generates fully synthetic patient data without relying on real EHR data, unlike data-driven models such as Syntegra \citep{syntegra2022technology}, MedGAN \citep{DBLP:journals/corr/ChoiBMDSS17}, and CTGan \citep{s23125644}. 
Synthetic data generated by rule-based data generators is preferable to data-driven models, as it allows the user to gain full control over disease trajectories and demographic parameters, thereby avoiding any biases introduced by underlying real-world data distributions.
Therefore, Synthea is well-suited to systematically evaluate the performance limits of various ML models for EHR-based disease classification.
The concrete contributions are as follows:
\begin{enumerate}
    \item Generation of patients with common diseases and a subgroup of simulated RD patients with a definable degree of dissimilarity for benchmarking purposes.
    \item No-code interface that reduces the barrier to entry for synthetic data generation.
    \item BMI-driven comorbidity modeling based on empirical evidence.
    \item Legacy module migration support.
\end{enumerate}


\section{Synthea Overview}
To fully understand SYNRARE, it is necessary to explain Synthea, as SYNRARE depends on its data generation capabilities.
Synthea generates a medical history of synthetic patients on predefined modules.
These modules typically represent a disease and generate several events, including symptom onset, encounters, medications, and other clinical events.
These events are also known as states.
This process simulates the patient's progression from the disease's initial manifestation to diagnosis, eventual recovery, or death.
For example, a module representing Cystic Fibrosis is a rare genetic disorder resulting in abnormally thick and sticky mucus clogging up the lungs and pancreas \citep{Ong2023CysticFA}.
The majority of affected patients are diagnosed early in their age through newborn screening, while there are a few receiving a diagnosis later in their age.
A sweat test is a reliable procedure for measuring the chloride level, where greater than or equal to 60 mmol/L indicates Cystic Fibrosis \citep{Ong2023CysticFA, Farrell2017DiagnosisOC}.
A state where the patient receives a diagnosis can therefore be defined as a probability exponential distribution for all patients, followed by another state capturing the chloride level in a uniform distribution between 60 - 80 mmol/L.
A synthetic dataset is generated based on the selected module for use in research studies.
This example is simplified and serves solely to illustrate the process of synthetic patient generation.
It is possible to manually edit modules to generate a rarer variant of a disease. 
However, this process demands a thorough understanding of Synthea's framework, domain knowledge, and the state being modeled.
Therefore, SYNRARE was developed to accelerate the generation of rarer disease variants and, therefore, enable them to establish a controlled test bed for research purposes.

\section{Features, Implementation and Results}\label{sec2}
One of the main features of Synthea is to view and modify an existing module.
However, a key missing component is applying a global modification to the module, meaning the current user must manually modify each state to accommodate their needs.

\subsection{Installation}
SYNRARE requires Python 3.8+ and Java JDK 11 to run both the software and Synthea.
Additional instructions on installing and running the software are available on GitLab.
After installing the software, the user can generate synthetic data based on the modules provided by Synthea.

\subsection{Module Selection}
SYNRARE offers three modes: Single Module, Multiple Modules, and All Modules.
Single Module allows the user to select, view, and modify states of the selected module.
The viewing and modification of the states are described later in this section.
Multiple Modules enable simultaneous patient generation by selecting a subset of available modules, thereby supporting the creation of customized disease combinations.
All Modules include all available modules and offer a quick way to generate data of patients representing the general population in a hospital.
The user can load their custom module into SYNRARE to either generate synthetic data or view and modify it.

\subsection{Module Viewing}
\begin{figure}
    \centering
    \includegraphics[width=1\linewidth]{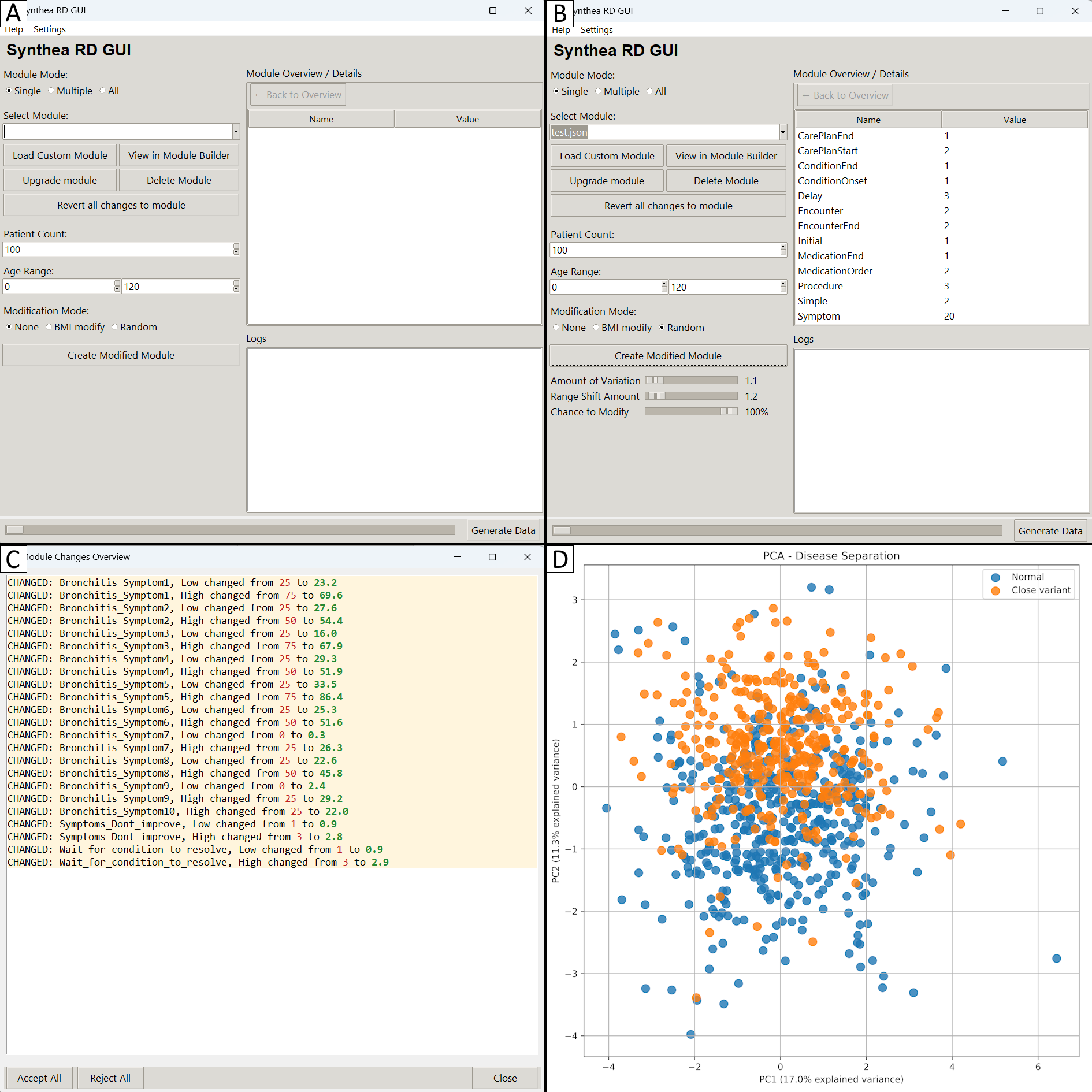}
    \caption{Overview of the SYNRARE GUI and a PCA plot. (A) shows the display upon startup. (B) is the detailed overview of the module "bronchitis". The user is able to modify each state independently or globally by adjusting the variance, range shift, and the probability to apply the change for each state. (C) represents the proposed change after applying a global modification. (D) is the PCA representation of the original "bronchitis" module and the modified module. }
    \label{fig:figure_1}
\end{figure}

SYNRARE extends Synthea's functionality by enabling users to view and modify the underlying distribution of each state in Single Module mode, as well as the probabilities associated with these events and their respective distribution types.
To be precise, these states follow a Uniform, Gaussian, or Exponential distribution or can be assigned an exact value.
The parameters for the Uniform distribution are set to span the range from the lower to the upper bound.
The Gaussian distribution requires both the mean and the variance.
Finally, the Exponential distribution only requires the mean.
A detailed overview of the module is shown in Figure \ref{fig:figure_1}A and \ref{fig:figure_1}B, allowing the user to check and modify some states.

Many legacy modules often contain missing fields or lack support for specific distributions. 
SYNRARE supports legacy module migration, allowing modules to be upgraded to the latest version and, therefore, enabling users to modify them accordingly. 

\subsection{Module Modification}
The user can apply the Random Modification to adjust the magnitude of the variance, the range shift, and the probability of the modification occurring, thereby globally changing each state as depicted in Figure \ref{fig:figure_1}B and \ref{fig:figure_1}C.
The modification depends on the distribution.
As the variance increases, the Uniform and Gaussian distributions become taller or wider, respectively.
In contrast, the Exponential distribution remains unaffected by the change.
A range shift moves the distribution, whereas a Uniform distribution shifts both its lower and upper bounds simultaneously to the left or right.
Additionally, the means of the Gaussian and Exponential distributions are similarly shifted, moving left or right as a result of this modification.
To illustrate the magnitude of the shift, if the user sets the range shift to 2, a state with a uniform distribution in the range of 25 to 50 could, in extreme cases, be shifted to the range of 50 to 75.
Therefore, the user should typically set the slide adjustment from 1 to 1.5 to apply minor modifications.

Another component of the module modification feature is the ability to adjust body mass index (BMI) to simulate known health effects across the categories: Normal, Overweight, Obesity Class I, Class II, and Class III.
SYNRARE automatically updates the associated health effects.
An individual with obesity shows increased risks of cardiovascular events, particularly in cases of severe obesity.
Epidemiological meta-analyses indicate that overweight adults (BMI $25–29.9$) have roughly 20–50\% higher odds of coronary heart disease and stroke, whereas obesity (BMI $\geq$ 30) roughly doubles the risk of these outcomes \citep{Guh2009TheIO}.
Furthermore, Table \ref{table_1} presents clinical evidence that the patient generally experiences more severe symptoms with increasing BMI.
For instance, the systolic blood pressure of an obesity III patient is modeled as being, on average, 25 mmHg higher than that of a normal-weight patient to simulate the high prevalence of hypertension in that group.
After applying any of the changes, the user receives a summary of the affected states and their updated values, which can be approved or rejected.
A new module reflecting the modifications is created; however, if the user is dissatisfied with the current modifications, they may remove it.
\begin{table*}[h] 
\begin{tabular}{
p{0.32\textwidth}
>{\centering\arraybackslash}p{0.15\textwidth}
>{\centering\arraybackslash}p{0.15\textwidth}
>{\centering\arraybackslash}p{0.15\textwidth}
>{\centering\arraybackslash}p{0.15\textwidth}
}
\hline
\textbf{Adjustment (unit)}   & \textbf{Overweight}    & \textbf{Obesity I}     & \textbf{Obesity II }   & \textbf{Obesity III} \\ \hline
Systolic/Diastolic BP increment (mmHg)   & $+5$          & $+10$         & $+15$         & $+25$                            \\
Resting heart rate increment (beats/min) & $+2$          & $+4$          & $+5$          & $+7$                             \\
Body temperature increment (°C)          & $+0.1$        & $+0.2$        & $+0.3$        & $+0.5$                           \\
Respiratory rate increment (breaths/min) & $+1$          & $+2$          & $+3$          & $+4$                             \\
Fasting glucose level multiplier         & $\times 1.05$ & $\times 1.10$ & $\times 1.15$ & $\times 1.24$                    \\
Total/LDL cholesterol multiplier         & $\times 1.05$ & $\times 1.15$ & $\times 1.25$ & $\times 1.40$                    \\
Liver enzymes (ALT, AST) multipler       & $\times 1.10$ & $\times 1.20$ & $\times 1.30$ & $\times 1.60$                    \\
General symptom severity multiplier      & $\times 1.10$ & $\times 1.20$ & $\times 1.30$ & $\times 1.50$                    \\
Disease probability multiplier           & $\times 1.15$ & $\times 1.35$ & $\times 1.65$ & $\times 2.10$                    \\ \hline
\end{tabular}
\caption{Parameter adjustment factors for each BMI category in the BMI modification module.}
\label{table_1}
\end{table*}

\subsection{Example case of generating a rare variant disease}
To exemplify the usability of SYNRARE, consider a use case involving bronchitis and a rare variant of lung disease for a classification task.
The user can modify the distribution for most states in the bronchitis module to create a close variant of the disease and thus determine the degree of similarity between bronchitis and the variant.
We selected the Random modification mode and set the variance, range shift, and probability to modify all states to 1.1, 1.2, and 100\%, respectively.
Thereafter, the user can apply different ML models to assess their performance on binary classification tasks.
Figure \ref{fig:figure_1}D depicts a PCA plot of bronchitis and its close variant.
Both diseases are similar, making simple, interpretable models, such as Logistic Regression, difficult to apply to binary classification tasks. One could consider instead whether applying complex black-box models yields stronger performance.
The user can create different scenarios to make the rare variant increasingly distinguishable from bronchitis to simulate classification tasks.
The use case is not limited to binary classification tasks; it can be extended to general supervised and unsupervised tasks, such as outlier detection and clustering.

\section{Discussion and Conclusion}
SYNRARE is an ideal solution for researchers seeking to develop and benchmark ML models for accelerating rare disease diagnosis.
An intuitive user interface enables domain experts to generate synthetic data and customize and tailor their modules for their test scenarios without compromising patient privacy, making SYNRARE an essential tool for supporting the rapid development and testing of ML approaches tailored for RD. 
SYNRARE enables users to generate synthetic data for their specific scenarios and subsequently rapidly benchmark ML algorithms used to assess their limitations.
This software is particularly suitable for users with domain knowledge to model and generate comparable disease variants.
To make the best use of SYNRARE, the user should ideally have domain knowledge to increase the realism of their disease scenario.
This includes, but is not limited to, a strong understanding of disease progression, familiarity with clinical guidelines, and knowledge of clinical phenotypes that characterize the disease.
Users should recognize the inherent limitations of Synthea, as synthetic data may not fully represent the complexity and variability present within real-world EHRs.
In practice, EHRs, particularly those of rare disease patients, often contain missing data and confounding information \citep{li_imputation_2021, ren_moving_2024}.
In contrast, Synthea generates "idealized" disease trajectories that include relevant symptoms and clinical interventions, but it nevertheless allows the systematic delineation of the capabilities of ML tools by creating datasets of varying difficulty.

\section{Acknowledgements}
We thank our former master's student, Andreas Rosenstjerne, for implementing and testing SYNRARE.

\section{Conflicts of interest}
Nothing declared.

\section{Funding}
This work was supported by the Innovative Medicines Initiative 2 Joint Undertaking (JU) under grant agreement No. 101034427. 

\bibliographystyle{oup-abbrvnat}
\bibliography{reference}

@article{Huda2021AML,
  title={A machine learning model for identifying patients at risk for wild-type transthyretin amyloid cardiomyopathy},
  author={Ahsan Huda and Adam Casta{\~n}o and Anindita Niyogi and Jennifer Schumacher and Michelle Meredyth Stewart and Marianna Bruno and Mo Hu and Faraz S. Ahmad and Rahul C. Deo and Sanjiv J. Shah},
  journal={Nature Communications},
  year={2021},
  volume={12}
}

@article{Cohen2020DetectingRD,
  title={Detecting rare diseases in electronic health records using machine learning and knowledge engineering: Case study of acute hepatic porphyria},
  author={Aaron M. Cohen and Steven Chamberlin and Thomas G. Deloughery and Michelle U. Nguyen and Steven Bedrick and Stephen Meninger and John J Ko and Jigar J Amin and Alex Wei and William R. Hersh},
  journal={PLoS ONE},
  year={2020},
  volume={15}
}

@article{Wilson2023DevelopmentOA,
  title={Development of a rare disease algorithm to identify persons at risk of Gaucher disease using electronic health records in the United States},
  author={Amanda Wilson and Alexandra Chiorean and Mario Aguiar and Davorka Sekuli{\'c} and Patrick Pavlick and Neha Shah and Lisa Sniderman King and Marie G{\'e}nin and Melissa Rollot and Margot Blanchon and Simon Gosset and Martin Montmerle and Cliona Molony and Alexandra Dumitriu},
  journal={Orphanet Journal of Rare Diseases},
  year={2023},
  volume={18}
}

@article{walonoski2018synthea,
  title={Synthea: An approach, method, and software mechanism for generating synthetic patients and the synthetic electronic health care record},
  author={Walonoski, Jason and Kramer, Mark and Nichols, Joseph and Quina, Andre and Moesel, Chris and Hall, Dylan and Duffett, Carlton and Dube, Kudakwashe and Gallagher, Thomas and McLachlan, Scott},
  journal={Journal of the American Medical Informatics Association},
  volume={25},
  number={3},
  pages={230--238},
  year={2018},
  publisher={Oxford Academic}
}

@article{Guh2009TheIO,
  title={The incidence of co-morbidities related to obesity and overweight: A systematic review and meta-analysis},
  author={Daphne Guh and Wei Zhang and Nick Bansback and Zubin Amarsi and Laird Birmingham and Aslam H. Anis},
  journal={BMC Public Health},
  year={2009},
  volume={9},
  pages={88 - 88},
}

@misc{syntegra2022technology,
  author       = {{Syntegra Inc.}},
  title        = {Synthetic Data Engine},
  year         = {2022},
  howpublished = {\url{https://www.syntegra.io/technology}},
  note         = {Accessed: 2026-02-25}
}

@article{DBLP:journals/corr/ChoiBMDSS17,
  author       = {Edward Choi and
                  Siddharth Biswal and
                  Bradley A. Malin and
                  Jon Duke and
                  Walter F. Stewart and
                  Jimeng Sun},
  title        = {Generating Multi-label Discrete Electronic Health Records using Generative
                  Adversarial Networks},
  journal      = {CoRR},
  volume       = {abs/1703.06490},
  year         = {2017},
  url          = {http://arxiv.org/abs/1703.06490},
  eprinttype    = {arXiv},
  eprint       = {1703.06490},
  bibsource    = {dblp computer science bibliography, https://dblp.org}
}

@Article{s23125644,
AUTHOR = {Alabsi, Basim Ahmad and Anbar, Mohammed and Rihan, Shaza Dawood Ahmed},
TITLE = {Conditional Tabular Generative Adversarial Based Intrusion Detection System for Detecting Ddos and Dos Attacks on the Internet of Things Networks},
JOURNAL = {Sensors},
VOLUME = {23},
YEAR = {2023},
NUMBER = {12},
ARTICLE-NUMBER = {5644},
URL = {https://www.mdpi.com/1424-8220/23/12/5644},
PubMedID = {37420810},
ISSN = {1424-8220},
DOI = {10.3390/s23125644}
}

@article{li_imputation_2021,
	title = {Imputation of missing values for electronic health record laboratory data},
	volume = {4},
	issn = {2398-6352},
	url = {https://doi.org/10.1038/s41746-021-00518-0},
	doi = {10.1038/s41746-021-00518-0},
	pages = {147},
	number = {1},
	journaltitle = {npj Digital Medicine},
	shortjournal = {npj Digital Medicine},
	author = {Li, Jiang and Yan, Xiaowei S. and Chaudhary, Durgesh and Avula, Venkatesh and Mudiganti, Satish and Husby, Hannah and Shahjouei, Shima and Afshar, Ardavan and Stewart, Walter F. and Yeasin, Mohammed and Zand, Ramin and Abedi, Vida},
	date = {2021-10-11},
}

@article{ren_moving_2024,
	title = {Moving Beyond Medical Statistics: A Systematic Review on Missing Data Handling in Electronic Health Records.},
	volume = {4},
	rights = {Copyright © 2024 C​opy​rig​h​t © 2024 Wenhui Ren et al.},
	issn = {2765-8783 2097-1095},
	doi = {10.34133/hds.0176},
	pages = {0176},
	journaltitle = {Health data science},
	shortjournal = {Health Data Sci},
	author = {Ren, Wenhui and Liu, Zheng and Wu, Yanqiu and Zhang, Zhilong and Hong, Shenda and Liu, Huixin},
	date = {2024},
	pmid = {39635227},
	pmcid = {PMC11615160},
	note = {Place: United States},
}

@article{Valdez2016PublicHA,
  title={Public Health and Rare Diseases: Oxymoron No More},
  author={Rodolfo Valdez and Lijing Ouyang and Julie C Bolen},
  journal={Preventing Chronic Disease},
  year={2016},
  volume={13},
}

@article{Health2024TheLF,
  title={The landscape for rare diseases in 2024.},
  author={The Lancet Global Health},
  journal={The Lancet. Global health},
  year={2024},
  volume={12 3},
  pages={
          e341
        },
}

@techreport{dubief2024newborn,
  author       = {Dubief, Jessie and Gross, Edith Sky and Faye, Fatoumata},
  title        = {Voices on Newborn Screening: The Opinion of People Living with a Rare Disease},
  institution  = {EURORDIS--Rare Diseases Europe},
  year         = {2024},
  month        = {May},
  type         = {Rare Barometer Survey Report},
  note         = {Conducted within the framework of the European Screen4Care research project (IMI2 Grant Agreement No. 101034427)},
  url          = {https://www.eurordis.org/wp-content/uploads/2024/05/RB_NBS_report_vff.pdf}
}

@article{Fecho2022LeveragingOE,
  title={Leveraging Open Electronic Health Record Data and Environmental Exposures Data to Derive Insights Into Rare Pulmonary Disease},
  author={Karamarie Fecho and Stanley C. Ahalt and Michael Knowles and Ashok K. Krishnamurthy and Margaret Leigh and Kenneth Morton and Emily R. Pfaff and Max Wang and Hong Yi},
  journal={Frontiers in Artificial Intelligence},
  year={2022},
  volume={5}
}

@article{Farrell2017DiagnosisOC,
  title={Diagnosis of Cystic Fibrosis: Consensus Guidelines from the Cystic Fibrosis Foundation},
  author={Philip M. Farrell and Terry B. White and Clement L. Ren and Sarah E. Hempstead and Frank Accurso and Nico Derichs and Michelle S. Howenstine and Susanna A. McColley and Michael J Rock and Margaret Rosenfeld and Isabelle Sermet-Gaudelus and Kevin W. Southern and Bruce C. Marshall and Patrick R. Sosnay},
  journal={The Journal of Pediatrics},
  year={2017},
  volume={181},
  pages={S4–S15.e1}
}

@article{Ong2023CysticFA,
  title={Cystic Fibrosis: A Review.},
  author={Thida Ong and Bonnie W Ramsey},
  journal={JAMA},
  year={2023},
  volume={329 21},
  pages={
          1859-1871
        },
  url={https://api.semanticscholar.org/CorpusID:259090508}
}

\end{document}